\documentclass[final]{cvpr}

\usepackage{times}
\usepackage{epsfig}
\usepackage{graphicx}
\usepackage{amsmath}
\usepackage{amssymb}
\usepackage{multirow}

\usepackage{pifont}% http://ctan.org/pkg/pifont
\newcommand{\cmark}{\ding{51}\xspace}%
\usepackage[pagebackref=true,breaklinks=true,colorlinks,bookmarks=false]{hyperref}

\def\cvprPaperID{****} % *** Enter the CVPR Paper ID here
\def\confYear{CVPR 2021}
\begin{document}

%%%%%%%%% TITLE
\title{An Empirical Study of Vehicle Re-Identification on the AI City Challenge}
\newcommand*{\affaddr}[1]{#1}
\newcommand*{\affmark}[1][*]{\textsuperscript{#1}}
\author{Hao Luo\affmark[1],  Weihua Chen\affmark[1], Xianzhe Xu\affmark[1], Jianyang Gu\affmark[1], Yuqi Zhang\affmark[1], Chong Liu\affmark[1] \\
Yiqi Jiang\affmark[1], Shuting He\affmark[1], Fan Wang\affmark[1],  Hao Li\affmark[1]\\
\affaddr{\affmark[]Machine Intelligence Technology Lab, Alibaba Group} \\
{\tt\small michuan.lh@alibaba-inc.com}
}

\maketitle
\pagestyle{empty}
\thispagestyle{empty} %%%%%%% Please uncomment it for the final version!!!!!

%%%%%%%%% ABSTRACT
\begin{abstract}
This paper introduces our solution for the Track2 in AI City Challenge 2021 (AICITY21). The Track2 is a vehicle re-identification (ReID) task with both the real-world data and synthetic data. We mainly focus on four points, \ie training data, unsupervised domain-adaptive (UDA) training, post-processing, model ensembling in this challenge. (1) Both cropping training data and using synthetic data can help the model learn more discriminative features. (2) Since there is a new scenario in the test set that dose not appear in the training set, UDA methods perform well in the challenge. (3) Post-processing techniques including re-ranking, image-to-track retrieval, inter-camera fusion, etc, significantly improve final performance. (4) We ensemble CNN-based models and transformer-based models which provide different representation diversity. With aforementioned techniques, our method finally achieves 0.7445 mAP score, yielding the first place in the competition. Codes are available at \url{https://github.com/michuanhaohao/AICITY2021_Track2_DMT}.
\end{abstract}

%%%%%%%%% BODY TEXT
\section{Introduction}

Vehicle ReID is an important computer-vision task which aims to identify the target vehicle in images or videos across different cameras, especially without knowing the license plate information. Vehicle ReID is important for intelligent transportation systems (ITS) of the smart city. For instance, the technology can track the trajectory of the target vehicle and detect traffic anomalies. Recently, most of works have been based on deep learning methods in vehicle ReID, and these methods have achieved great performance in vehicle ReID.

\begin{figure}[htb]
\centering
\includegraphics[width=0.85\linewidth]{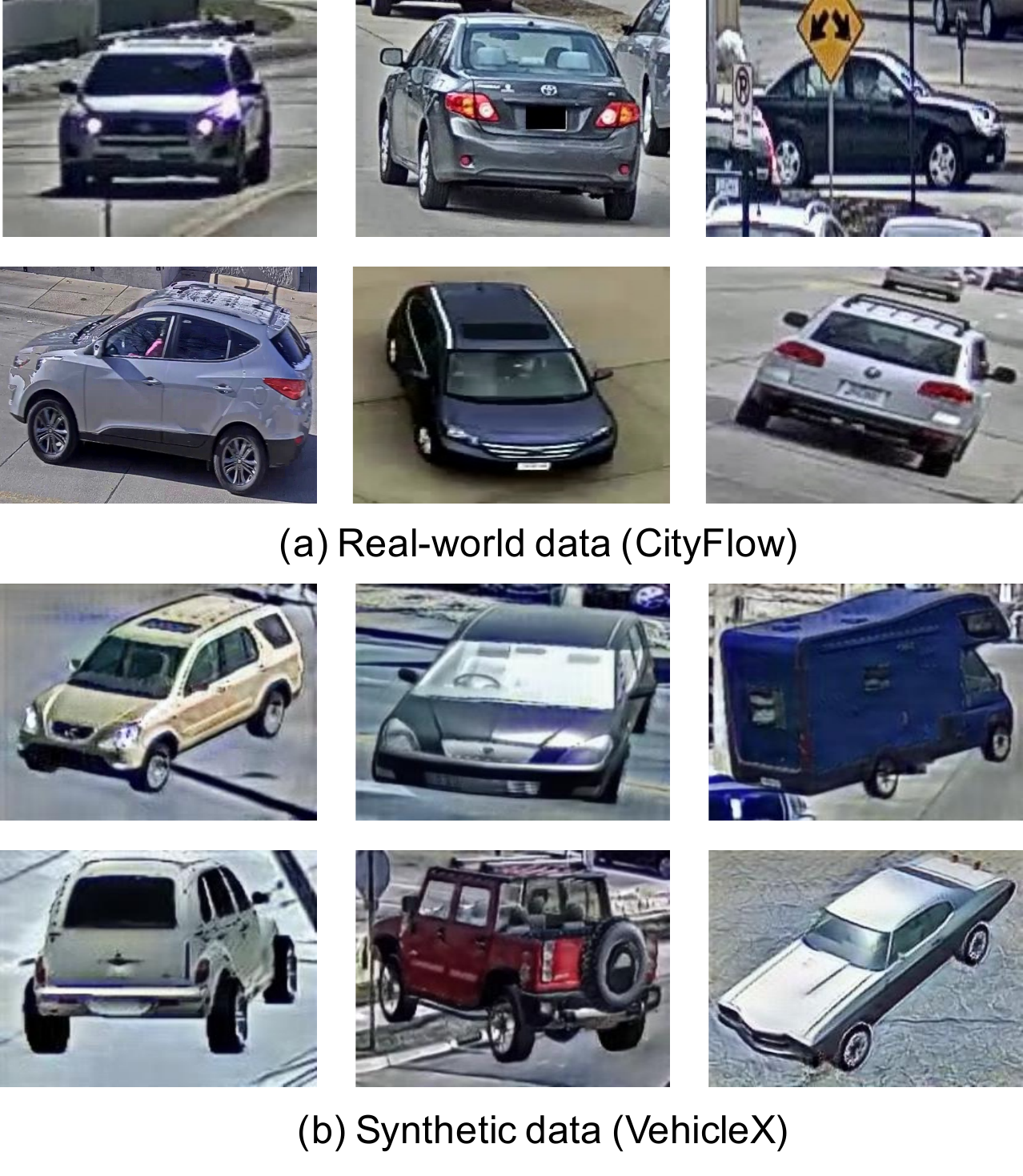}
\caption{Some examples of the real-world and synthetic data.}
\label{fig:demo}
\end{figure}

As shown in Figure \ref{fig:demo}, Track2 provides a real-world dataset CityFlow-V2 \cite{Tang19CityFlow} and a synthetic dataset VehicleX \cite{Yao19VehicleX} for model training. However, CityFlow-V2 only contains 52,712 images, which is not enough to train a robust model. Therefore, the first challenge is how to overcome the lack of real data. For the real data, inaccurate bounding boxes introduce noise in the image. We re-detect real-world images according to their heatmaps and add them into training data. Adding synthetic images into training data also can improve the performance. In addition, we also try to use SPGAN \cite{SPGAN} to transfer synthetic data into `real-world' data.

The second challenge observed is that a new scenario appears in the test set, \ie there exists domain bias between the training and test sets. Unsupervised domain-adaptive (UDA) methods are suitable to address the problem. In the second stage, We perform UDA methods to automatically generate pseudo labels on the testing data, which are used to fine-tune models. To improve the quality of pseudo-labels, we consider tracklet information and camera bias between images in the clustering process. The UDA training can significantly improve the model performance on the test set.

Thirdly, several post-processing techniques are leveraged during the inference stage. For example, a camera model and a orientation model are trained to reduce the camera and orientation bias between images, respectively. Since the tracklet information is provided, we integrate all features of a tracklet to obtain a more robust feature for the tracklet in the inference stage. Camera verification \cite{VehicleNet} based on a camera-connection constraint is also introduced to remove negative images from candidate images. In addition, some widely used methods including the re-ranking \cite{zhong2017re} and inter-camera fusion also improve a lot of accuracy. 

Finally, we ensemble multiple models to further improve scores on the leaderboard. It is note that we have try both CNN-based methods and transformer-based methods (\ie TransReID \cite{TransReID}). To our best knowledge, it is the first time to study pure-transformer models on the AI City Challenge. Our experience shows that TransReID can provide
diversity that is different from CNN-based models. Therefore, we ensemble CNN-based methods and TransReID to finally achieve 0.7445 mAP score, yielding the first place on the leaderboard. 

Our contributions can be summarized as follow:
\begin{itemize}
\item We conduct a empirical study of vehicle ReID on AICITY21. We have tried many methods that have been verified in the ReID field in the past few years. 
\item We observe domain bias between the training and testing data and introduce the UDA training to address the challenge.
\item We try transformer-based models on the AI City Challenge for the first time. The experimental results show that transformer-based models can provide representation diversity different from CNN-based models.
\item We achieves 0.7445 mAP socre without external data, yield the first place in the competition.
\end{itemize}

\section{Related Works}

We introduce deep ReID and some works of AICITY2020 in this section.

\subsection{Deep ReID}
Re-identification (ReID) is widely studied in the field of computer vision. This task possesses various important applications. Most existing ReID methods based on deep learning. Recently, CNN-based features have achieved great progress on both person ReID and vehicle ReID. Person ReID provides a lot of insights for vehicle ReID. Luo \etal \cite{luo2019bag, luo2019strong} proposed a strong baseline \cite{luo2019bag,luo2019strong} in person ReID, which also performers well in vehicle ReID. For vehicle ReID, Liu \emph{et al.} \cite{liu2016deep} introduced a pipeline that uses deep relative distance learning (DRDL) to project vehicle images into an Euclidean space, where the distance can directly measure the similarity of two vehicle images. Shen \emph{et al.} \cite{shen2017learning} proposed a two-stage framework that incorporates complex spatial-temporal information of vehicles to effectively regularize ReID results. Zhou \emph{et al.} \cite{zhou2018aware} designed a viewpoint-aware attentive multi-view inference (VAMI) model that only requires visual information to solve multi-view vehicle ReID problems. While He \emph{et al.} \cite{he2019part} proposed a simple yet efficient part-regularized discriminative feature-preserving method, which enhances the perceptive capability of subtle discrepancies, and reported promising improvement. Some works \cite{zhu2019vehicle,qian2019stripe} also studied discriminative part-level features for better performance. Some works \cite{wang2017orientation,khorramshahi2019attention,khorramshahi2019dual} in vehicle ReID had utilized vehicle key points to learn local region features. Several recent works \cite{he2019part, wang2019vehicle,guo2019two,teng2018scan} in vehicle ReID had stated that specific parts such as windscreen, lights and vehicle brand tend to have much discriminative information. In recently, a pure-transformer method called as TransReID \cite{TransReID} has shown that transformer-based methods achieves better performance than CNN-based methods on some vehicle ReID benchmarks.

\subsection{AICITY20}
Since AICITY21 is updated from AI CITY Challenge 2020 (AICITY20), some methods of AICITY20 are helpful for our solution. The organizers outlined the methods of leading teams in \cite{naphade20204th}. Zheng \etal \cite{zheng2020going} trained real data with synthetic data by applying style transformation and content manipulation. Zhu \etal \cite{zhu2020voc} proposed an approach named VOC-ReID, taking the triplet vehicle-orientation- camera as a whole and reforming background/shape similarity as camera/orientation re-identification. He \etal \cite{he2020multi} proposed the Identity Mining method to automatically generate pseudo labels on the test data to expand the training set. Some other works also studied on loss functions, model structures, post-processing strategies, etc.

\begin{figure*}[htb]
\centering
\includegraphics[width=0.9\linewidth]{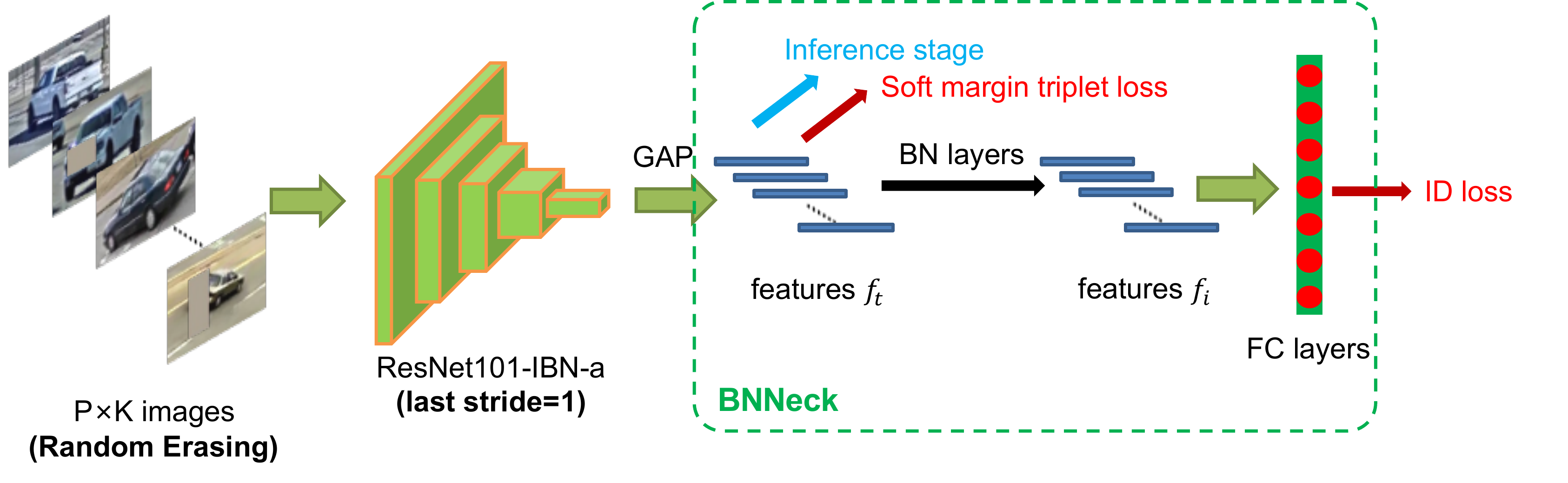}
\caption{The framework of our CNN-based baseline BoT \cite{he2020multi}.}
\label{fig:baseline}
\end{figure*}

\begin{figure*}[htb]
\centering
\includegraphics[width=0.9\linewidth]{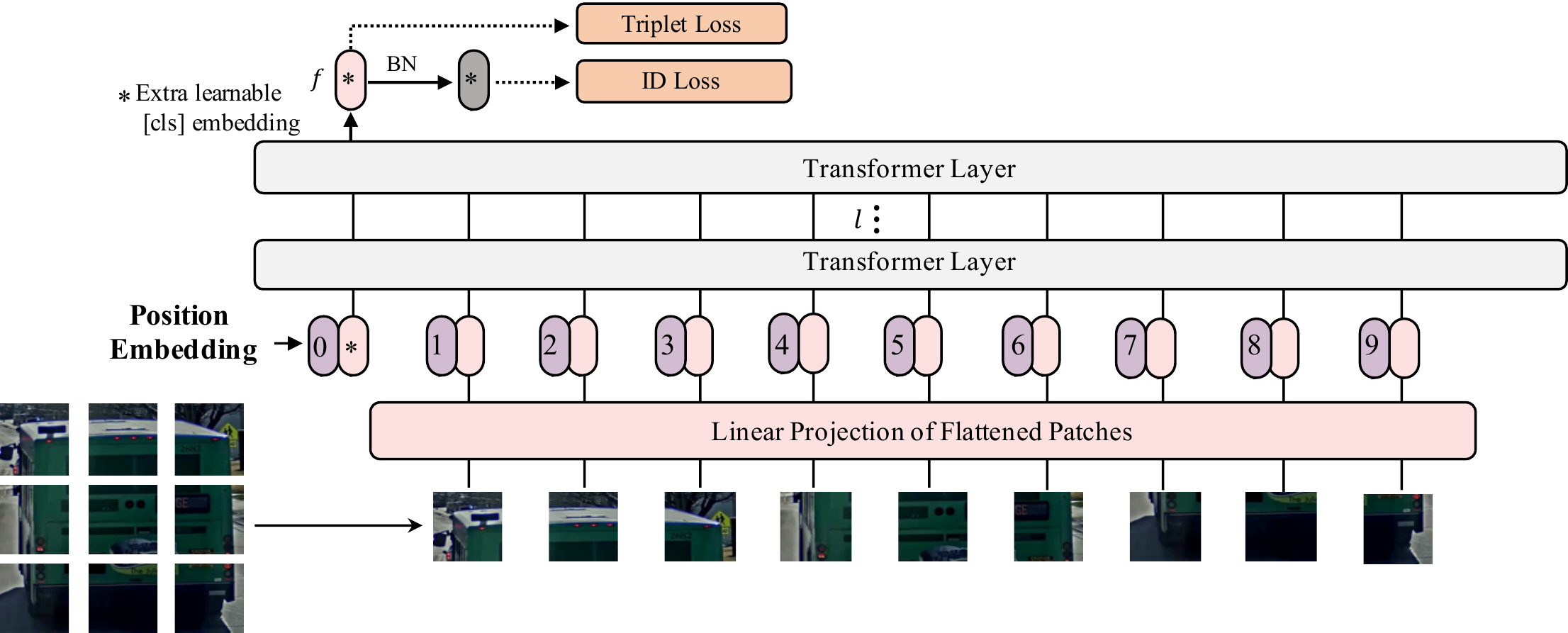}
\caption{The framework of our transformer-based baseline TransReID \cite{TransReID}.}
\label{fig:baseline2}
\end{figure*}

\section{Our Method}

Our solution includes three parts, \ie baseline training, UDA training, and post-processing.

\subsection{Stage1: Baseline Training}

In this section, we will introduce models and training data used to train robust baseline models.

\subsubsection{Baseline Model}

Baseline model is important for the final ranking. In track2, we use a CNN-based baseline \cite{luo2019bag,luo2019strong} shown in Figure \ref{fig:baseline} and a transformer-based baseline \cite{TransReID} shown in Figure \ref{fig:baseline2}. 

Similar with many past works on the AI CITY Challenge, we choose Bag of Tricks (BoT) as the CNN-based baseline. The version of BoT is the one proposed by ours \cite{he2020multi} in the 2020 AI City Challenge. In the modified version, both label smoothing and center loss are removed. In addition, the triplet loss is the soft-margin version as follow:
\begin{equation}
\mathcal{L}_{Tri}=\log \left[1+ \exp(||f_a -f_p||_2^2 - ||f_a -f_n||_2^2 +m)\right]
\end{equation}

We also use a transformer-based baseline called TransReID \cite{TransReID} that is the first pure-transform model in the field of ReID. To simplify our codes, we remove the local branch (the JPM branch) and only use the global feature of TransReID. But our experiments show that the JPM branch still performs well on CityFlow-V2. Since the dataset does not provide accurate orientation labels, we also ignore the SIE module. 

\subsubsection{Training data}
The Track2 provides two training sets (CityFlow-V2 and VehicleX). CityFlow-V2 is not a large-scale dataset to train robust ReID models. Therefore, a challenge is to overcome the lack of training data. VehicleX additionally provides 192,150 images to expand training data. However, all these images are synthesis images generated by a 3D engine. The domain bias between CityFlow-V2 and VehicleX should be considered. To address these challenges, we introduce our training sets as follow.

\begin{figure}[htb]
\centering
\includegraphics[width=0.9\linewidth]{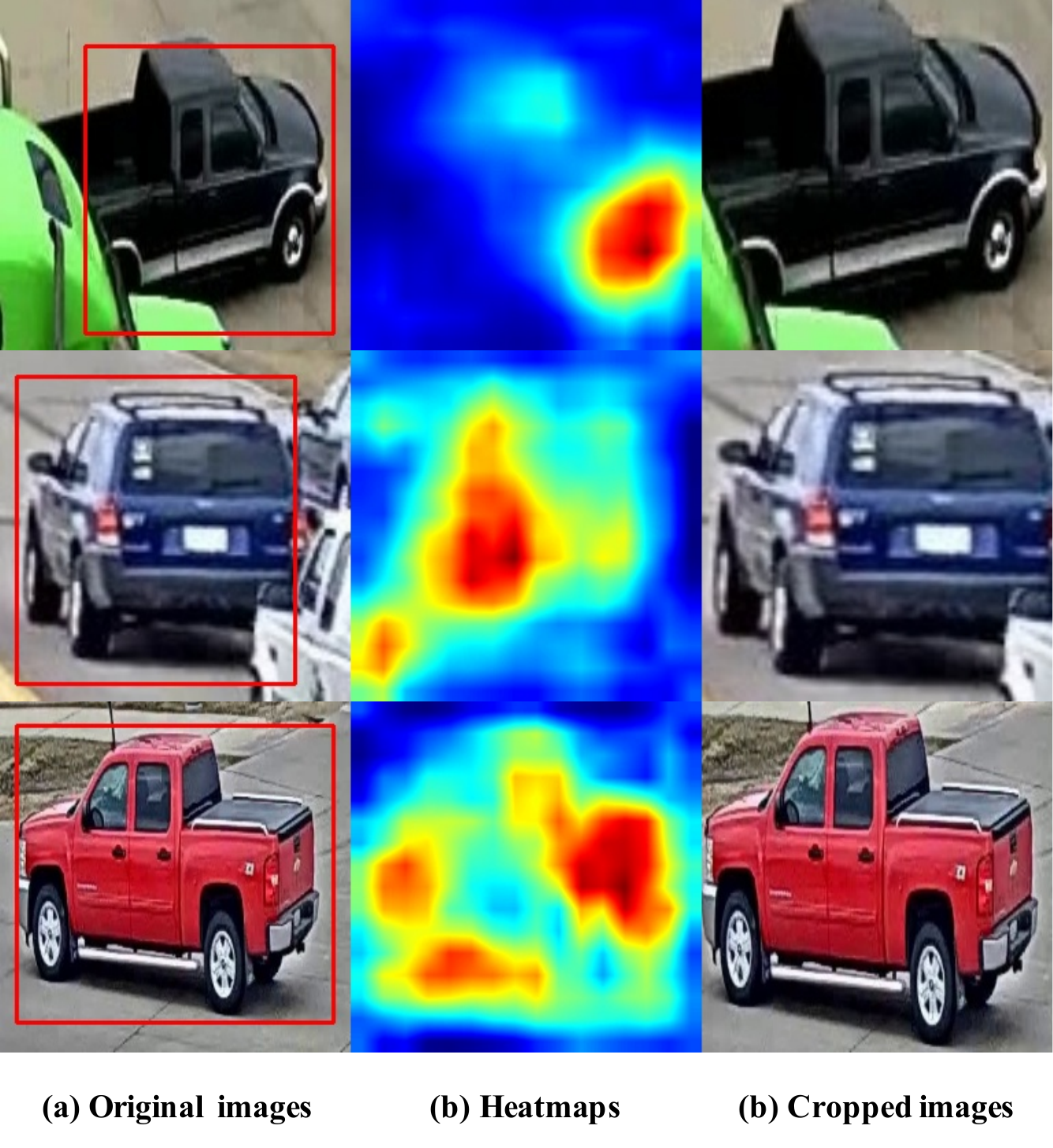}
\caption{Some examples of cropped images generated by weakly supervised detection.}
\label{fig:demo1}
\end{figure}

\begin{figure}[htb]
\centering
\includegraphics[width=1.0\linewidth]{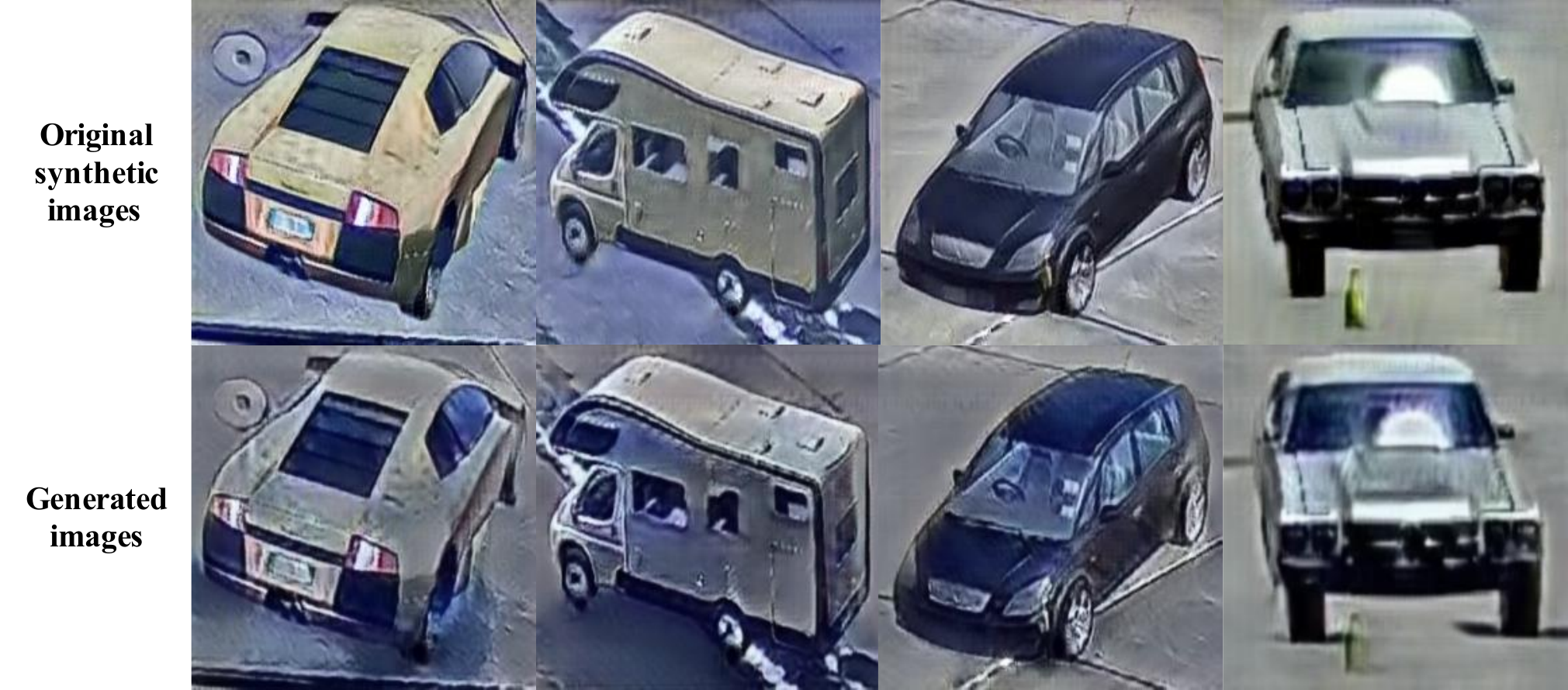}
\caption{Some examples of `real-world' generated by SPGAN. The first and second rows show original synthetic images and generated images, respectively.}
\label{fig:demo2}
\end{figure}

\textbf{CityFlow-V2 (CFV2).} The CityFlow-V2 dataset\cite{Naphade19AIC19,Tang19CityFlow} is a real-world dataset that is captured by 46 cameras in real-world traffic environment. It totally includes 85,058 images of 880 vehicles. 52,717 images of 440 vehicles are used for training. The remaining 31,238 images of 440 vehicles are for testing. It is note that the training set is captured by 40 cameras. In the test set, part of images are captured by 6 new cameras that do not appear in the training set.

\textbf{CityFlow-V2-CROP (CFV2-C).} To overcome the lack of real-world data, we use the weakly supervised detection \cite{zhu2020voc} to increase the training data. We train an initial vehicle ReID model to get heatmap response of each image and setting a threshold to get the bounding box larger than it. Then, we get a cropped copy of both training and test set. The modified dataset is called as CityFlow-V2-CROP (CFV2-C) in this paper. After weakly supervised detection, the real-world data becomes doubled. Some examples are shown in Figure \ref{fig:demo1}.

\textbf{VehicleX (VeX).} The VehicleX dataset is a synthetic dataset generated by a publicly available 3D engine VehicleX \cite{Yao19VehicleX}. The dataset only provides training set that contains 192,150 images of 1,362 vehicles in total. In addition, the attribute labels, such as car colors, car types, orientation labels are also annotated. In Track2, the synthetic data can be used for the model training or transfer learning. However, there exists domain bias between the real-world and synthetic data.

\textbf{VehicleX-SPGAN (VeX-S).} We use SPGAN \cite{SPGAN} to transfer synthetic images to `real-world' images to reduce the domain bias. The generated images construct a new dataset called as VehicleX-SPGAN (VeX-S) in this paper. The number of images in VehicleX-SPGAN is same with the number of images in VehicleX.

\subsection{Stage2: UDA Training}

As we mentioned above, a new scenario appears in the test set, which results in the domain bias between the training and test sets. It is a typical UDA ReID task. In the second stage, we use clustering algorithms to generate pseudo labels on the testing data and then fine-tune baseline models. 

For an image $I$ that belongs to the $C$-th camera and the $T$-th tracklet in the test set, its global feature is denoted as $g_I$. Then we compute the average feature of all images captured by the $C$-th camera and denote the average feature as $\overline{g}_C$. To reduce the camera bias between cross-camera images, the single-frame feature $f_I$ of $I$ is computed as follow:
\begin{equation}
f_I= g_I - \alpha \overline{g}_C \quad,
\end{equation}\label{ICF}
where $\alpha$ is the balance weight between $g_I$ and $\overline{g}_C$. When we average features of all IDs in the $c$-th camera, the ID information is ignored. Therefore, $\overline{g}_C$ can represent the camera information without requirement of training an extra camera model on the test set. We also try to train a camera model on the training set, but it performs worse than computing $\overline{g}_C$. 

In addition, since the tracklet ID of each image is provided, we can integrate all features of the tracklet $T$ to get the tracklet feature $t_I$ of $I$. Finally, the final feature of $I$ is computed as follow:
\begin{equation}
\hat{f}_I= \beta f_I + (1-\beta) t_I \quad ,
\end{equation}
where $\beta$ is the balance weight between single-frame and tracklet features. $\hat{f}_I$ is fed into a cluster to generate pseudo labels. Similar with previous works \cite{ge2020mutual,gu20201st}, DBSCAN is chosen as the cluster in this paper. Then, we fine-tune baseline models trained in the Stage1. In the second stage (Stage2), all models are trained for only 3 epochs with a small learning rate. $\beta$ is set to $0.0005$ for the UDA training.

\subsection{Post-Processing}
Post-processing can significantly improve ReID performance in the inference stage. In this section, we will introduce several post-processing methods in this paper.

\textbf{Augmentation Test.} For each test image, we extract features of original image (CVF2) and cropped image (CFV2-C). Then, we flip these two images horizontally and additionally extract two features. We get four features totally and then average them to get the final ReID feature. 

\textbf{Re-Ranking.} We adopt a widely used Re-ranking (RK) method \cite{zhong2017re} to update the final result. we set $k1=7$, $k2=2$, $\lambda=0.6$ in this paper.

\textbf{Weighted Tracklet Features.} The tracklet IDs are provided for the testing set in Track2. A prior knowledge is that all frames of a tracklet belong to the same ID. In the inference stage, standard ReID task is an image-to-image (I2I) problem. However, with the tracklet information, the task becomes an image-to-track (I2T) problem. For the I2T problem, the feature of tracklet is represented by features of all frames of the tracklet. He \etal \cite{he2019multi} compared average features (AF) and weighted features (WF) of tracklets. In this paper, we calculated the weighted features for each tracklet.

\textbf{Camera Verification.} Zheng \etal \cite{VehicleNet} applied the assumption that the query image and the target images are taken in different cameras.  Given a query image, we remove the images of the same camera ID from candidate images.

\textbf{Inter-Camera Fusion.} As we introduce in Eq \ref{ICF}, we also update a ReID feature with its inter-camera feature in the inference stage. The camera bias can be reduced. $\beta$ is set to $0.18$ for the post-processing.

\textbf{Camera and Orientation Bias.} Inspired by \cite{VehicleNet, zhu2020voc}, we train a camera model and a orientation model on CityFlow-V2 and VehicleX, respectively. Because VechcleX has orientation label available, we split the angle (0 - 360) into 36 bins, each bin is treated as an orientation ID. After training camera and orientation models, we can calculate camera and orientation similarities between an image pair. The fusion distance matrix can be expressed as following \cite{zhu2020voc}:
\begin{equation}
D = D_r - \lambda_1 D_c - \lambda_2 D_o, \quad ,
\end{equation}
where $D_r$, $D_c$ and $D_o$ are ID distance matrix, orientation distance matrix and camera distance matrix, respectively. In this paper, we set $\lambda_1 =0.1$ and $\lambda_2 =0.05$.

\section{Experiments}

\subsection{Implementation Details}

\textbf{CNN-Based Models in Stage1.} All the images are resized to 384 $\times$ 384. As for data augmentation, we use random flipping, random padding and random erasing. In the training stage, we use soft margin triplet loss with the mini-batch of 8 identities and 12 images of each identity which leads to better convergence. SGD is adopted as the optimizer and the initial learning rate is set to $1e^{-2}$. Besides, we adopt the Warmup learning strategy and spend 10 epochs linearly increasing the learning rate from $1e^{-3}$ to $1e^{-2}$. The learning rate is decayed to $1e^{-3}$ and $1e^{-4}$ at 40th and 70th epoch, respectively. We totally train the model with 80 epochs. We adopt ResNet-IBN \cite{he2016deep,pan2018IBN-Net}, DenseNet-IBN \cite{huang2017densely,pan2018IBN-Net}, ResNest \cite{zhang2020resnest}, SeResNet-IBN \cite{pan2018IBN-Net} and ResNext-IBN \cite{pan2018IBN-Net} as backbones. All backbone are pre-trained on ImageNet \cite{deng2009imagenet}.

\textbf{Transformer-Based Models in Stage1.} All the images are resized to 256 $\times$ 256 because of the limited GPU memory. The augmentation, batch size and optimizer are same with CNN-based models. However, we adopt cosine learning rate with only 40 epochs for TransReID because it converges faster than CNN models. The backbone is ViT/16-Base \cite{VIT} pre-trained on ImageNet. In order to simplify our codes in this paper, the JPM and SIE modules are not used.

\textbf{UDA Training in Stage2.} In the stage2, the DBSCAN is adopted as the cluster. All models are fine-tuned for only three epochs. The learning rate is fixed to $4e^{-4}$. To increase diversity, each model is fine-tuned with two sets of pseudo labels generated by two different cluster parameters. 

\subsection{Validation Data}
Since each team has only 20 submissions, it is necessary to use the validation set to evaluate methods offline. We split the training set of CityFlow-V2 into the training set and the validation set. For convenience, the validation set is denoted as Split-Test. Split-Test includes 18701 images of 88 vehicles. 

\subsection{The Ablation Study of Training Data}  
\renewcommand{\multirowsetup}{\centering}
\begin{table}[htb]
    \begin{center}\small
    \begin{tabular}{c|cccc|cc}
    \hline
    Datasets & CVF2 & CVF2-C & VeX & VeX-S & mAP & R-1 \\
    \hline
     Type1 & \cmark &  &  &   &36.0  &51.7  \\ 
     Type2 & \cmark &  &\cmark  &   &46.2   &64.8  \\ 
     Type3 & \cmark &\cmark  &\cmark  &   &49.1  &65.9  \\ 
     Type4 & \cmark &\cmark  &  &\cmark   &49.9  &66.4  \\ 
    \hline
    \end{tabular}
    \end{center}
    \caption{\label{tab:data} The performance of models trained with different training sets on Split-Test. The backbone is ResNet50-IBN-a and all training images are resized to 256 $\times$ 256.}
\end{table}

To explore the effectiveness of training data, we compare models trained with different training sets in Table \ref{tab:data}. We train a ResNet50-IBN-a model with images resized to 256 $\times$ 256 as the baseline. When we only use CVF2 to train the model, it achieves 36.0\% mAP on Split-Test. The synthetic dataset VeX improves the performance by a large margin, which achieves 46.2\% mAP on Slpit-Test. In addition, the cropped data CVF2-C brings 2.9\% mAP gains because of relatively loose cropping in CityFlow-V2. However, the `real-world' data VeX-S generated by SPGAN does not surpass VeX by a lot. We infer that it may be because we did not adjust the parameters of SPGAN on the CityFlow-V2, so the quality of the generated data is not very high. 

\subsection{Comparison of Different Backbones}  

\renewcommand{\multirowsetup}{\centering}
\begin{table}[htb]
    \begin{center}
    \begin{tabular}{c|c|cc}
    \hline
    Datasets & Backbones & mAP & R-1 \\
    \hline
    \hline
     Type1 & ResNet50-IBN-a &36.0  &51.7  \\ 
     Type1 & ResNet101-IBN-a&38.8  &54.8  \\ 
     Type1 & TransReID      &42.1  &59.8  \\ 
     \hline
     Type2 & ResNet50-IBN-a &46.2   &64.8  \\ 
     Type2 & ResNet101-IBN-a&47.6   &68.2  \\ 
     Type2 & TransReID      &45.5   &61.0  \\ 
    \hline
    \end{tabular}
    \end{center}
    \caption{\label{tab:backbones} The performance of different backbones is compared on Split-Test. All training images are resized to 256 $\times$ 256. Type1 means we use ony CityFlow-V2 to train models. Type2 means we use both CityFlow-V2 and VehicleX to train models.}
\end{table}

We comparison two CNN-based backbones (\ie ResNet50-IBN-a and ResNet101-IBN-a) and a transformer-based backbone (\ie TransReID) in Table \ref{tab:backbones}. For a fair comparison, all training images are resized to 256 $\times$ 256. 

When we only use CityFlow-V2 to train models, TransReID achieves better performance than ResNet50-IBN-a and ResNet101-IBN-a, which shows that the transformer-based models has a strong representation ability. However, when the synthetic dataset VehicleX is added into training, TransReID obtains the worst performance in these three models. The reason may be that TransReID is easier to overfit to the training data than CNN-based backbones.

\subsection{The Ablation Study of Two-Stage Training} 

To evaluate the effectiveness of UDA training in the Stage2, we conduct experiments on the ResNet50-IBN-a baseline in the Table \ref{tab:two}. DBSCAN clusters all test images into approximately 900 IDs. After UDA training, the performance of the model increases from 46.2\% to 66.0\% mAP. Due to the limited time, we did not verify the influence of inter-camera fusion and tracklet features too much. But both of them can significantly improve quality of pseudo labels, and then improve the performance of the model. Another experience we have observed is that fine-tuning the model too many epochs will harm the performance of the model.

\begin{table}[htb]
    \begin{center}
    \begin{tabular}{l|cc}
    \hline
    Methods & mAP & R-1 \\
    \hline
    \hline
     Stage1  &46.2   &64.8  \\ 
     Stage2  &66.0  &76.8  \\ 
    \hline
    \end{tabular}
    \end{center}
    \caption{\label{tab:two} The ablation study of two-stage training on Split-Test. The backbone is ResNet50-IBN-a.}
\end{table}

\subsection{The Ablation Study of Post-Processing} 

We show the ablation study of post-processing in Table \ref{tab:post}. The baseline is ResNet50-IBN-a trained on CityFlow-V2 and VehicleX. These post-processing methods improve the performance by almost 30\% mAP totally on Split-Test, which shows the effectiveness of post-processing methods.

\renewcommand{\multirowsetup}{\centering}
\begin{table}[tb]
    \begin{center}
    \begin{tabular}{l|cc}
    \hline
    Methods & mAP & R-1 \\
    \hline
    \hline
     Baseline  &46.2   &64.8  \\ 
     + Augmentation Test    &47.0  &65.0  \\ 
     + Re-Ranking           &58.8  &66.1  \\ 
     + Weighted Tracklet Features  &66.8  &66.9  \\ 
     + Camera Verification  & 72.7 &73.3  \\ 
     + Inter-Camera Fusion  & 74.7 &75.4  \\ 
     + Camera and Orientation Bias  & 76.0  & 75.9  \\ 
    \hline
    \end{tabular}
    \end{center}
    \caption{\label{tab:post} The ablation study of post-processing methods on Split-Test. With all post-processing methods, ResNet50-IBN-a achieves 0.6055 mAP score on the leaderboard.  }
\end{table}

\subsection{Ablation Study of the Solution}

We present the ablation study of different parts in Table \ref{tab:ablation}. Most of results have been present in aforementioned tables. In overview, training data, uda training, post-processing improve the performance by about 10\%, 20\%, 30\% mAP on Split-Test, respectively. However, post-processing methods only improve 4.3\% mAP and 5.0\% Rank-1 accuracies on Split-Test after the UDA training. For reference, `Baseline (Type3) + Post-processing' achieves 0.6055 mAP score. The UDA training improve the score to 0.6665 on the leadeborad.

\renewcommand{\multirowsetup}{\centering}
\begin{table}[tb]
    \begin{center}
    \begin{tabular}{l|cc}
    \hline
    Methods & mAP & R-1 \\
    \hline
    \hline
     Baseline (Type1)  &36.0  &51.7  \\ 
     Baseline (Type3)  &46.2   &64.8  \\ 
     + UDA Training  &66.0  &76.8  \\
     + Post-processing  & 76.0  & 75.9  \\ 
     + UDA Training \& Post-processing & 80.3  & 80.9  \\ 
    \hline
    \end{tabular}
    \end{center}
    \caption{\label{tab:ablation} The ablation study of all parts on Split-Test. With the UDA training and post-processing methods, ResNet50-IBN-a achieves 0.6665 mAP score on the leaderboard.}
\end{table}

\renewcommand{\multirowsetup}{\centering}
\begin{table}[tb]\small
    \begin{center}
    \setlength\tabcolsep{4.5pt}
    \begin{tabular}{l|cc|cc|cc}
    \hline
              & \multicolumn{2}{c|}{Stage1}  & \multicolumn{2}{c|}{Stage2} & \multicolumn{2}{c}{Post}\\
    Backbones & mAP & R-1 & mAP & R-1 & mAP & R-1\\
    \hline
    \hline
     ResNet101-IBN-a  &54.0   &72.3 &71.2 &79.2 &81.6 &83.5  \\ 
     ResNet101-IBN-a$^{\dagger}$  &54.9   &72.7 &72.5 &80.2 &82.2 &83.5  \\ 
     ResNet101-IBN-a$^{\ddagger}$  &54.3   &71.5 &71.4 &79.0 &82.6 &84.6  \\ 
     ResNext101-IBN-a  &50.6   &70.2 &74.0 &82.5 &82.5 &84.0  \\ 
     ResNest101  & 51.2 & 70.8 & 72.4 & 80.8 & 82.5 & 84.1 \\
     SeResNet101-IBN-a & 51.6 & 69.9 & 73.1 & 81.6 & 82.1 & 84.3 \\
     DenseNet169-IBN-a & 51.2 & 69.2 & 69.9 & 79.6 & 82.7 & 85.9 \\
     TransReID & 47.5 & 65.2 & 52.4 & 70.3 & 78.5 & 81.3 \\
     \hline
     Ensemble (CNN) & - & - &- &- &84.3 &85.9\\
     Ensemble (All) & - & - &- &- &84.8 &86.7\\
    \hline
    \end{tabular}
    \end{center}
    \caption{\label{tab:ensemble} Detailed results of single models and model ensemble on Split-Test. `Post' means Post-Processing. $^{\dagger}$ means the training data is Type4 (\ie CFV2 + CFV2-C + VeX-S). $^{\ddagger}$ means the augmentation is a little different. `Ensemble (CNN)' means we ensemble 7 CNN models apart from TransReID. `Ensemble (All)' means we ensemble all 8 models. ResNext101-IBN-a achieves 0.7058 mAP score on the leaderboard.}
\end{table}

\subsection{Model Ensemble} 

We ensemble several models to boost the performance on the leaderboard. ResNet101-IBN-a, DenseNet169-INB-a, ResNext101-IBN-a, SeResNet101-IBN-a, ResNest101 and TransReID are adopted as backbones. Unless otherwise specified, the training data is set to Type3 (\ie CFV2 + CFV2-C + VeX) in the Stage1. With different training settings, we train 8 models in the Stage1. We then adopt the UDA training to fine-tine these models in the Stage2. Detailed results are present in Table \ref{tab:ensemble}. An interesting phenomena we observed is that TransReID can provide representation diversity different from CNN models. When we ensemble 7 CNN models, Ensemble (CNN) achieves 84.3\% mAP and 85.9\% rank-1 accuracy on Split-Test. However, when we continue to integrate TransReID, the performance increases to 84.8\% mAP and 86.7\% rank-1 accuracy even that TransReID achieves only 78.5\% mAP and 81.3\% rank-1 accuracy. Therefore, the diversity of TransReID plays an important role in the model ensemble. 

It is note that we try two different cluster parameters for each model in the Stage2. Thus we train totally 16 models. In our experience, the performance of a single model ranges from 0.69 to 0.72 mAP scores. For instance, ResNext101-IBN-a achieves 0.7058 mAP score. We ensemble these 16 models, and achieves 0.7445 mAP score on the final leaderboard, yielding the first place in the Track2.

\subsection{Competition Results}

Our team (Team ID 47) achieves 0.7445 in the mAP score which achieves the first place in the 2021 NVIDIA AI City Challenge Track 2. As shown in the Table \ref{tab:leaderboard}, it is the performance of top-10 teams. Our codes are available at \url{https://github.com/michuanhaohao/AICITY2021_Track2_DMT}.

\begin{table}[tb]
  \begin{center}
  \begin{tabular}{c|c|c|c}
\hline
    Rank &Team ID	& Team Name & mAP Scores 	 \\
 	\hline
	\hline
    1	& \textbf{47}	& \textbf{DMT (Ours)}	& \textbf{0.7445} \\
    2	&9	    &NewGeneration	&0.7151\\
    3	&7	    &CyberHu	    &0.6650\\
    4	&35	    &For Azeroth	&0.6555\\
    5	&125	&IDo	        &0.6373\\
    6	&44	    &KeepMoving	    &0.6364\\
    7	&122	&MegVideo	    &0.6252\\
    8	&71	    &aiem2021	    &0.6216\\
    9	&61	    &CybercoreAI	&0.6134\\
    10	&27	    &Janus Wars	    &0.6083\\
    \hline
  \end{tabular}
  \end{center}
  \caption{\label{tab:leaderboard}Competition results of AICITY21 Track2.}
\end{table}

\section{Conclusion}

In this paper, we conduct an empirical study of vehicle ReID on the 2021 AI City Challenge. We verify the UDA training is important in this challenge. In addition, transformer-based models are first time to be studied in the AI City Challenge. We believe transformer-based methods have great potential for ReID tasks. Finally, our solution yield the first place in the Track2 of the 2021 AI City Challenge.

{\small
\bibliographystyle{ieee_fullname}
\bibliography{egbib}
}

\end{document}